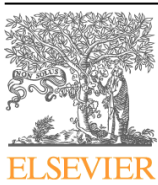

# Master-Auxiliary: an efficient aggregation strategy for video anomaly detection

Zhiguo Wang[a], *, Zhongliang Yang[a] and Yujin Zhang [a]

[a] *Image Engineering Laboratory, Tsinghua University, Beijing, 100084, China*

## ABSTRACT

The aim of surveillance video anomaly detection is to detect events that rarely or never happened in a certain scene. Generally, different detectors can detect different anomalies. This paper proposes an efficient strategy to aggregate multiple detectors. First, the aggregation strategy chooses one detector as master detector by experience, and sets the remaining detectors as auxiliary detectors. Then, the aggregation strategy extracts credible information from auxiliary detectors, including credible abnormal (Cred-a) frames and credible normal (Cred-n) frames. After that, the frequencies that each video frame being judged as Cred-a and Cred-n are counted. Applying the events' time continuity property, more Cred-a and Cred-n frames can be inferred. Finally, the aggregation strategy utilizes the Cred-a and Cred-n frequencies to vote to calculate soft weights, and uses the soft weights to assist the master detector. Experiments are carried out on multiple datasets. Comparing with existing aggregation strategies, the proposed strategy achieves state-of-the-art performance.

---

* Corresponding author. e-mail: wzg16@mails.tsinghua.edu.cn

# 1. Introduction

Surveillance videos play an important role in safety protection. However, it is time-consuming and labor-intensive for people to watch long hours of surveillance videos. To address this problem, automatic video anomaly detection is needed. Detecting video anomalies is a challenging task, because the kinds of abnormal events are unbounded, moreover, it is infeasible to list all kinds of possible abnormal events and gather enough samples for each kind.

Many efforts have contributed to this task [1–6]. Generally, different detectors can detect different anomalies. Better performance can be expected by aggregating multiple detectors together. The existing aggregation strategies can be classified into three categories: weighted sum strategy [7–11], competition strategy [12–14], and cascade strategy [15–17].

Weighted sum strategy [7–11] calculates the weighted sum of multiple anomaly scores to detect anomalies. A drawback of this strategy is that it does not consider the credibility of the fused information, which results in that it combines both the correct information and error information of multiple detectors. The performance of this strategy trends to achieve the average performance of multiple detectors.

Competition strategy [12][13,14] utilizes the maximum or the minimum anomaly scores of a single frame to judge anomaly. It is usually used to aggregate detectors belonging to the same kind, and is hard to aggregate different kinds of detectors. The reason is that different kinds of detectors have different anomaly thresholds, the anomaly scores belonging different kinds of detectors are not comparable.

The cascade strategy [15–17] uses multiple detectors in cascade. They detect and discard strong normal samples in the former detectors, and transmit the remaining samples to the next detector. It exhibits well performance. The shortcoming of this strategy is that: once a misjudgment occurs in the former detectors, the subsequent detectors cannot correct this error.

In this paper, we propose a new aggregation strategy: Master-Auxiliary Aggregation Strategy (MAAS). The pipeline of MAAS is shown in Fig.1.

Firstly, MAAS chooses one detector as master detector by experience, and sets the remaining detectors as auxiliary detectors.

Then, with the help of the training data, MAAS automatically calculates a Cred-a threshold and a Cred-n threshold for each auxiliary detector, and utilizes the thresholds to detect Cred-a and Cred-n frames. After that, MAAS counts the Cred-a and Cred-n frequencies for each frame and applies the events' time continuity property to infer more Cred-a and Cred-n frames: the frames between two time-neighbored Cred-a frames can be inferred as Cred-a frames, the frames between two time-neighbored Cred-n frames can be inferred as Cred-n frames.

Finally, MAAS utilizes the Cred-a and Cred-n frequencies of a frame to vote to calculate soft weight, and uses soft weights to refine the master detector.

The proposed MAAS has the following advantages: (i) It extracts credible information from auxiliary detectors to assist the master detector. In this way, it reduces the unreliable information that fused into the aggregated detector. (ii) It does not compare anomaly scores belonging to different detectors. Therefore, it can aggregate different kinds of detectors conveniently. (iii) It extracts both Cred-a and Cred-n frames from auxiliary detectors, and applies time continuity property to infer more credible frames. Therefore, comparing with the cascade strategy, MAAS extracts more credible information. (iv) It utilizes a voting mechanism to use credible information, that makes the credible information more robust. (v) It adopts a soft weight method to fuse the credible information to the master detector. It is more efficient than believing the credible frames without any doubt.

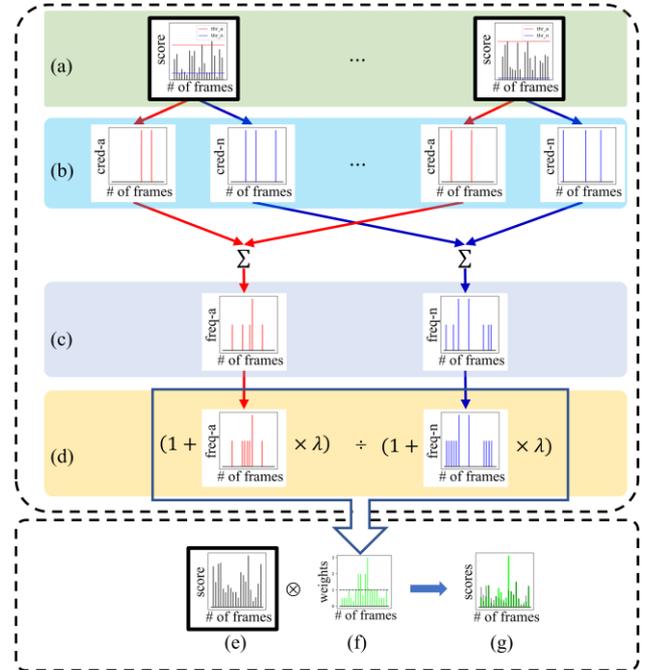

**Fig. 1.** The pipeline of the proposed aggregation strategy. (a) Anomaly scores in auxiliary detectors. (b) The detected Cred-a and Cred-n frames in auxiliary detectors. (c) Cred-a frequencies (freq-a) and Cred-n frequencies (freq-n) of frames. (d) The frequency information after time continuity process. The formula indicates the voting mechanism to calculate soft weights. (e) The master detector. (f) The soft weights calculated by (d). (g) The aggregated detector. The green lines indicate the refined anomaly scores. The black lines indicate the raw anomaly scores in master detector.

Experiments are carried out on the UCSD Ped2 dataset and the Avenue dataset. Comparing with the existing aggregation strategies, the proposed strategy achieves state-of-the-art performance.

# 2. Related work

In this section, the related works are introduced from two aspects: (1) Existing anomaly detection models. (2) Existing aggregation strategies.

## 2.1. Existing anomaly detection models

Anomaly detection models can be roughly divided into two categories: traditional machine learning models [1,3] and deep learning models[4,5].

(1) Traditional machine learning models use hand-crafted-features or deep-features to construct feature space, then utilize the traditional machine learning methods to learn the distribution of the feature space and to detect anomalies. Mahadevan et al. [18,19] applied mixture of dynamic textures (MDT) model to detect temporal and spatial anomalies. Kratz et al.[20] coupled multiple hidden Markov models (HMMs) to detect anomalies. Hu et al. [21] applied topic model to detect anomalies. Hinami et al. [22] employed pretrained convolutional neural network (CNN) to extract deep features and used the one-class support vector machine (OC-SVM) to detect anomalies. Yuan et al. [23] utilized sparse coding to detect anomalies. They firstly learned an over-complete dictionary to reconstruct normal samples, then judged the samples with large reconstruction errors as anomalies.

(2) Deep learning models use neural networks to learn the manifold distribution of normal samples and judge the samples that deviate from this distribution as abnormal. Fan et al. [24] utilized Gaussian mixture variational autoencoder (VAE) to classify samples into multiple clusters and utilized the conditional probabilities of the test samples to detect anomalies. Tang et al. [25] utilized generative adversarial network (GAN) to detect anomalies. They firstly train a U-net and a discriminator network in an adversarial way, and then judge the samples with low discriminator probabilities as anomalies. Many works [26][27] utilize generation error (GE) to detect anomalies. They firstly utilize GNN to generate normal samples, then judge the samples with large GE as anomalies. Hasan et al. [26] employed auto-encoder (AE) to reconstruct normal frames, and judged the samples with high reconstruction errors as anomalies. Liu et al. [27] applied U-net to predict future frames, and adopted the prediction error to detect anomalies.

*2.2. Existing aggregation strategies.*

The existing aggregation algorithms can be divided into three categories: weighted sum strategy [7–11], competition strategy [12–14], cascade strategy [15–17].

(1) Weighted-sum strategy [7–11]. Luo et al. [7] employed the weighted sum of the two GE anomaly scores to detect anomalies. Lee et al.[9] utilized the weighted sum of the GE-based anomaly scores and the discriminator anomaly scores to detect anomalies. Ravanbakhsh et al. [11] trained two GANs. They added the two score maps generated by two discriminators to detect anomalies.

(2) Competition strategy [12–14]. Wang et al. [12] utilized multiple HMMs to detect anomalies. They judged a sample as anomaly once one of the HMMs judged it as an anomaly. Sabokrou et al. [13] applied two Gaussian detectors to detect anomalies. They classified a sample as anomaly if both detectors recognized it. Bao et al. [14] ensembled multiple AE detectors. A sample is judged as an anomaly when all detectors identified it.

(3) Cascade strategy [15–17]. Sabokrou et al. [16] cascaded multiple Gaussian detectors to detect anomalies. Wang et al. [17] cascaded two VAE networks to detect anomalies. They first employed a shallow VAE to detect and reduce unnecessary normal samples. Then, they utilized the deep VAE to detect anomalies in remaining samples.

As discussed in the first section, these strategies have different shortcomings. This paper proposes a new strategy to solve these problems.

## 3. The proposed method

This section first introduces a model which can generate multiple detectors at the same time, then introduces the proposed aggregation strategy in detail.

*3.1. Model*

In deep learning methods, the GE-based methods achieved very good performance. Following the work [27], this paper utilizes U-Net as GNN to generate future frames with multiple losses, include intensity loss, gradient loss, flow loss and discriminator loss. Different from work [27], in the test phase, we use multiple losses to generate multiple detectors, rather than using only one of them. The architecture of the model is shown in Fig. 2. The details of the U-Net network and discriminator network are similar to that in [27].

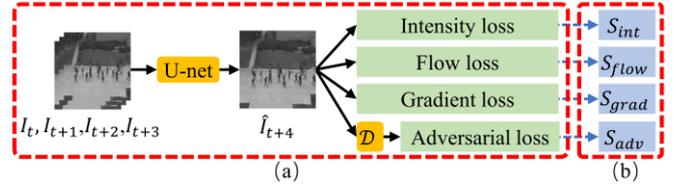

**Fig. 2.** The architecture of the model which can generate multiple detectors. (a) The architecture of the model and multiple losses in the training period. $\mathcal{D}$ indicates a discriminator. (b) Multiple detectors in the testing period.

Let $\mathcal{G}$ be the U-Net, $\mathcal{D}$ be the discriminator, $\hat{I}$ be the output of $\mathcal{G}$, and $I$ be the ground truth of $\hat{I}$. The intensity loss map $GE_{map}^{int}$, the gradient loss map $GE_{map}^{grad}$, the flow loss map $GE_{map}^{flow}$ can be calculated as follows:

$$GE_{map}(I, \hat{I}) = \sum_c |\hat{I}_{i,j} - I_{i,j}|_1 \quad (1)$$

$$GE_{map}^{int}(I_t, \hat{I}_t) = (GE_{map}(I_t, \hat{I}_t))^2 \quad (2)$$

$$GE_{map}^{grad}(I_t, \hat{I}_t) = GE_{map}(|\hat{I}_{t,i,j} - \hat{I}_{t,i-1,j}|, |I_{t,i,j} - I_{t,i-1,j}|) \\ + GE_{map}(|\hat{I}_{t,i,j} - \hat{I}_{t,i,j-1}|, |I_{t,i,j} - I_{t,i,j-1}|) \quad (3)$$

$$GE_{map}^{flow}(I_t, \hat{I}_t) = GE_{map}(\mathcal{F}(\hat{I}_t, I_{t-1}), \mathcal{F}(I_t, I_{t-1})) \quad (4)$$

where $c$ indicates the channels of the feature map, $i,j$ denote the spatial index in the frame; $\mathcal{F}$ represents the Flow-net [28] which is used to generate optical flow maps.

Utilizing the GE maps and the discriminator $\mathcal{D}$, the losses in the training period can be represented as:

$$L_{int}(I, \hat{I}) = \sum_{i,j} GE_{map}^{int}(I_t, \hat{I}_t) \quad (5)$$

$$L_{grad}(I, \hat{I}) = \sum_{i,j} GE_{map}^{grad}(I_t, \hat{I}_t) \quad (6)$$

$$L_{flow}(I, \hat{I}) = \sum_{i,j} GE_{map}^{flow}(I_t, \hat{I}_t) \quad (7)$$

$$L_{adv}^{\mathcal{G}}(\hat{I}) = \frac{1}{2}\sum_{i,j}(\mathcal{D}(\hat{I})_{i,j} - 1)^2 \quad (8)$$

$$L_{adv}^{\mathcal{D}}(I, \hat{I}) = \sum_{i,j}\frac{1}{2}(\mathcal{D}(I)_{i,j} - 1)^2 + \sum_{i,j}\frac{1}{2}(\mathcal{D}(\hat{I})_{i,j})^2 \quad (9)$$

$$L_{\mathcal{G}} = \omega_{int}L_{int}(I, \hat{I}) + \omega_{grad}L_{grad}(I, \hat{I}) \\ + \omega_{flow}L_{flow}(I, \hat{I}) + \omega_{adv}L_{adv}^{\mathcal{G}}(\hat{I}) \quad (10)$$

$$L_{\mathcal{D}} = L_{adv}^{\mathcal{D}}(I, \hat{I}) \quad (11)$$

where $i,j$ denote the spatial index of the feature map, $L_{\mathcal{G}}$ is the loss function for training $\mathcal{G}$, $L_{\mathcal{D}}$ is the loss function for training $\mathcal{D}$.

In the testing period, following work [29], the block-level GEs are used to calculate anomaly scores.

$$S_{int}(t) = max\left(mean_{30}(GE_{map}^{int}(I_t, \hat{I}_t))\right) \quad (12)$$

$$S_{grad}(t) = max\left(mean_{30}(GE_{map}^{grad}(I_t, \hat{I}_t))\right) \quad (13)$$

$$S_{flow}(t) = max\left(mean_{30}(GE_{map}^{flow}(I_t, \hat{I}_t))\right) \quad (14)$$

where $mean_{30}$ indicates the 2D mean blur operation with filter size 30. The anomaly score generated by $\mathcal{D}$ is represented as:

$$S_{adv}(t) = max(1 - \mathcal{D}(I_t)) \quad (15)$$

Therefore, this model can generate 4 detectors at the same time: $S_{int}, S_{grad}, S_{flow}, S_{adv}$.

*3.2. Aggregation strategy*

Let $\{S_1, ..., S_i, ..., S_D\}$ be $D$ anomaly detectors, where $S_i = [s_i(1), ..., s_i(t), ..., s_i(T^{test})]$ is the $i$-th detector, $s_i(t)$ is the anomaly score for frame $I_t$, $T^{test}$ is the number of the frames of a video.

The MAAS first chooses one detector as master detector, denoted as $S^{master}$, and sets the remaining detectors as auxiliary detectors, denoted as $\bar{\bar{S}}_1, ..., \bar{\bar{S}}_i, ..., \bar{\bar{S}}_{D-1}$.

Assuming that the higher the anomaly score, the higher the confidence that the sample is abnormal; the lower the anomaly score, the higher the confidence that the sample is normal, MAAS calculates two thresholds automatically for each auxiliary detector to detect Cred-a and Cred-n frames.

$$thr_i^a = \bar{\bar{S}}_{i,train}^{max_{\alpha*T_{train}}} * \gamma_a \quad (16)$$

$$thr_i^n = \bar{\bar{S}}_{i,train}^{min_{\beta*T_{train}}} * \gamma_n \quad (17)$$

where $thr_i^a$ is the threshold calculated to detect Cred-a frames in $\bar{\bar{S}}_i$, $thr_i^n$ is the threshold calculated to detect Cred-n frames in $\bar{\bar{S}}_i$; $\bar{\bar{S}}_{i,train}^{max_{\alpha*T_{train}}}$ indicates the $(\alpha * T_{train})$-th largest anomaly score in the training data in $\bar{\bar{S}}_i$, $\bar{\bar{S}}_{i,train}^{min_{\beta*T_{train}}}$ represents the $(\beta * T_{train})$-th smallest anomaly score in the training data in $\bar{\bar{S}}_i$; $T_{train}$ represents the number of frames in training dataset; $\alpha$ indicates the false-alarm-rate in the training dataset; $\beta$ indicates the strong-normal-rate in the training dataset. $\gamma_a$ and $\gamma_n$ are the strict coefficients, where $\gamma_a > 1$ and $\gamma_n < 1$. The higher the $\gamma_a$, the stricter the $thr_i^a$ is set; the lower the $\gamma_n$, the stricter the $thr_i^n$ is set.

It is noticeable that, in order to set $thr_i^a$ and $thr_i^n$ automatically, we introduced 4 hyperparameters: $\alpha, \beta, \gamma_a, \gamma_n$. All these hyperparameters can be set according to human's prior knowledge or application needs.

Utilizing $thr_i^a$ and $thr_i^n$, the Cred-a and Cred-n frames can be detected in $\bar{\bar{S}}_i$:

$$C_i^a(t) = \begin{cases} 1, & s_i(t) \geq thr_i^a \\ 0, & otherwise \end{cases} \quad (18)$$

$$C_i^n(t) = \begin{cases} 1, & s_i(t) \leq thr_i^n \\ 0, & otherwise \end{cases} \quad (19)$$

where $C_i^a(t)$ indicates whether $I_t$ is judged as Cred-a frame or not in $\bar{\bar{S}}_i$, $C_i^n(t)$ indicates whether $I_t$ is judged as Cred-n or not in $\bar{\bar{S}}_i$.

Then, MAAS counts the Cred-a frequency $f_a(t)$ and Cred-n frequency $f_n(t)$ for each frame $I_t$:

$$f_a(t) = \sum_{i=1}^{D-1} C_i^a(t) \quad (20)$$

$$f_n(t) = \sum_{i=1}^{D-1} C_i^n(t) \quad (21)$$

The occurrence of events is continuous in time. The time-neighbored Cred-a frames or Cred-n frames can be considered belonging to the same abnormal event or normal event. Therefore, applying this property and using the detected Cred-a and Cred-n frames, more credible frames can be inferred:

$$f_a'(t) = max\left(min(f_a(t_1), f_a(t_2)), f_a(t)\right), \\ if\ t_1 < t < t_2, t_2 - t_1 \leq \varepsilon_a \quad (22)$$

$$f_n'(t) = max(min(f_n(t_1), f_n(t_2)), f_n(t)), \\ if\ t_1 < t < t_2, t_2 - t_1 \leq \varepsilon_n \quad (23)$$

where $f_a'(t)$ and $f_n'(t)$ are the inferred Cred-a and Cred-n frequencies; $\varepsilon_a$ and $\varepsilon_n$ represent the minimum number of consecutive frames of an abnormal event and a normal event. Equation (22) means that, if $I_{t_1}$ and $I_{t_2}$ are two time-neighbored Cred-a frames, the inner frame $I_t$ between them should also be a Cred-a frame, the Cred-a frequency of $I_t$ should at least equal to the minimum value of $f_a(t_1)$ and $f_a(t_2)$. In practice, $\varepsilon_n$ can be smaller than $\varepsilon_a$, because sometimes the duration of abnormal events is short. It is easy to miss the short-term anomalous event if the $\varepsilon_n$ is set too long.

Finally, MAAS employs $f_a'(t)$ and $f_n'(t)$ to calculate soft weights via a voting manner as shown in equation (24), then utilizes the soft weights to assist the $S^{master}$, as shown in equation (25):

$$\omega_{soft}(t) = (1 + f_a'(t) * \lambda)/(1 + f_n'(t) * \lambda) \quad (24)$$

$$\tilde{S}(t) = S^{master}(t) * \omega_{soft}(t) \quad (25)$$

where $\omega_{soft}(t)$ is the soft weight calculated for frame $I_t$; $\tilde{S}(t)$ is the final anomaly score in the aggregated detector for $I_t$; $\lambda \geq 0$ is a hyperparameter which indicates the degree of applying the credible information to assist the master detector. If $\lambda = 0$, it indicates that the MAAS does not use auxiliary detectors to assist master detector.

*3.3. summary*

Algorithm 1 describes the aforementioned steps as a summary.

---
**Algorithm 1**

**Input**: multiple detectors $\{(S_{1,train}, S_{1,test}), ..., (S_{D,train}, S_{D,test})\}$.
**Hyperparameters**: $\alpha, \beta, \gamma_a, \gamma_n, \varepsilon_a, \varepsilon_n, \lambda$
**Output**: Aggregated detector $\tilde{S}$
1: choose one detector as $S^{master}$, the others as auxiliary detectors.
$\{(S_{train}^{master}, S_{test}^{master})\}, \{(\bar{\bar{S}}_{1,train}, \bar{\bar{S}}_{1,test}), ..., (\bar{\bar{S}}_{D-1,train}, \bar{\bar{S}}_{D-1,test})\}$
2: for $i$ in range $(1, D-1)$:
$\quad thr_i^a = \bar{\bar{S}}_{i,train}^{max_{\alpha*T_{train}}} * \gamma_a$
$\quad thr_i^n = \bar{\bar{S}}_{i,train}^{min_{\beta*T_{train}}} * \gamma_n$
$\quad C_i^a = \begin{cases} 1, & \bar{\bar{S}}_{i,test} \geq thr_i^a \\ 0, & otherwise \end{cases}$
$\quad C_i^n = \begin{cases} 1, & \bar{\bar{S}}_{i,test} \leq thr_i^n \\ 0, & otherwise \end{cases}$
3: $f_a = \sum_{i=1}^{D-1} C_i^a$, $f_n = \sum_{i=1}^{D-1} C_i^n$
4: for $t$ in range $(1, T_{test})$:
$\quad f_a'(t) = max\left(min(f_a(t_1), f_a(t_2)), f_a(t)\right),$
$\quad\quad t_1 < t < t_2, t_2 - t_1 \leq \varepsilon_a$
$\quad f_n'(t) = max\left(min(f_n(t_1), f_n(t_2)), f_n(t)\right),$
$\quad\quad if\ t_1 < t < t_2, t_2 - t_1 \leq \varepsilon_n$
5: $\omega_{soft}(t) = (1 + f_a'(t) * \lambda)/(1 + f_n'(t) * \lambda)$
6: $\tilde{S}(t) = S_{test}^{master}(t) * \omega_{soft}(t)$
7: return $\tilde{S}$

---

In this strategy, there are 7 hyperparameters: $\alpha, \beta, \gamma_a, \gamma_n, \varepsilon_a, \varepsilon_n, \lambda$. As mentioned in section 3.2, they all have clear application meanings and can be set according to the application needs and human's prior knowledge.

**4. Experiments**

This section evaluates and analyzes MAAS on two publicly available benchmark datasets: the CUHK Avenue dataset [30] and the UCSD Pedestrian dataset [18].

*4.1. Datasets*

The UCSD Pedestrian dataset [18] and CUHK Avenue dataset [30] are two commonly used datasets in video anomaly detection

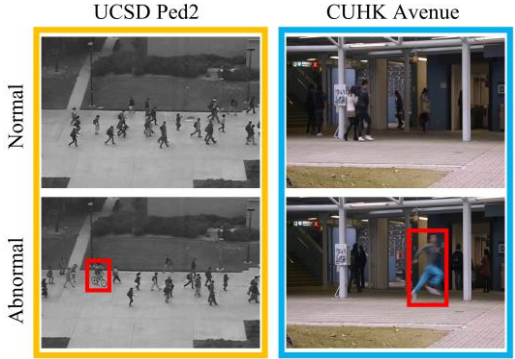

**Fig. 3.** Examples of normal and abnormal frames in UCSD Ped2 dataset and CUHK Avenue dataset. Red boxes denote anomalies in abnormal frames.

task. Some samples of the datasets are shown in Fig. 3.

**USCD dataset.** The UCSD Pedestrian dataset is composed of two subsets: Ped1 and Ped2. There are wrong labels in Ped1 subset [31]. Therefore, only Ped2 subset is used in this paper. Ped2 subset contains 16 training and 12 test videos. Anomalous events include cycling, skateboarding, crossing lawns, cars, etc. The frame resolution is $360 \times 240$ pixels.

**CUHK Avenue dataset.** It contains 16 training and 21 test videos. Anomalous events include running, throwing objects etc. The frame resolution is $360 \times 640$ pixels.

*4.2. Evaluation metric*

The most commonly used evaluation metric is the Receiver Operation Characteristic (ROC) curve and the Area Under this Curve (AUC). A higher AUC value indicates better anomaly detection performance. Following the work [11][29], the frame-level AUC is adopted to evaluate the anomaly detection performance.

*4.3. Implementation details*

In our experiments, all video frames are resized to $256 \times 256$, and pixel values are normalized to [-1, 1].

In the training process, the Adam [32] is adopted as optimizer and the mini-batch size is set to 4. For gray scale datasets, the learning rate of $\mathcal{G}$ and $\mathcal{D}$ are set as 0.0001 and 0.00001, respectively, while for color scale datasets, they are set as 0.0002 and 0.00002 respectively. $\omega_{int} = 1$, $\omega_{grad} = 1$, $\omega_{flow} = 2$, $\omega_{adv} = 0.05$.

In the testing period, anomaly scores of different detectors are calculated by equations (12)-(15). Anomaly scores in each detector are smoothed using a 1D median blur operation with median filter radius 15. The values of hyperparameters in the aggregation strategy are set as: $\alpha = 0.01, \beta = 0.1, \gamma_a = 2, \gamma_n = 0.99, \varepsilon_a = 80, \varepsilon_n = 40, \lambda = 10$. For $S_{adv}$, $\gamma_a$ is set as 1.01.

*4.4. Performance*

The performances of the aggregated detectors and their corresponding master detectors are shown in table 1. In table 1 and following sections, we use $\tilde{S}_{agg}^*$ represents the aggregated detector in which $S_*$ is set as $S^{master}$.

As shown in table 1, the performances of the aggregated detectors surpass the corresponding master detectors significantly. That shows the effectiveness of the proposed strategy.

**Table 1.** AUCs of the aggregated detectors with different detectors as master detectors. In each cell, the first value is the AUC of the master detector, the second value is the AUC of the corresponding aggregated detector.

| | Ped2 | Avenue |
|---|---|---|
| $\tilde{S}_{agg}^{int}$ | (0.9678, **0.9934**) | (0.8982, **0.9256**) |
| $\tilde{S}_{agg}^{flow}$ | (0.98, **0.9867**) | (0.8607, **0.905**) |
| $\tilde{S}_{agg}^{grad}$ | (0.9689, **0.9934**) | (0.842, **0.9121**) |
| $\tilde{S}_{agg}^{adv}$ | (0.5759, **0.9856**) | (0.8441, **0.8838**) |

*4.5. Comparing with existing aggregation strategies*

To demonstrate the efficiency of the proposed strategy, we compare the proposed strategy with the existing aggregation strategies:

**Weight-sum** $(\tilde{S}_w)$: using the weighted sum of multiple losses to detect anomalies. The weights are same to those in the training process.

$$\tilde{S}_w(t) = \omega_{int} S_{int}(t) + \omega_{grad} S_{grad}(t) + \omega_{flow} S_{flow}(t) + \omega_{dis} S_{adv}(t) \quad (26)$$

**Weight-sum-norm** $(\tilde{S}_{wn})$: using the weighted sum of multiple normalized anomaly scores to detect anomalies. As discussed in [29], the anomaly scores are normalized in the whole dataset.

$$S'(t) = \frac{S(t) - min(S)}{max(S) - min(S)} \quad (27)$$

$$\tilde{S}_{wn}(t) = S'_{int}(t) + S'_{grad}(t) + S'_{flow}(t) + S'_{adv}(t) \quad (28)$$

where $S'(t)$ is the normalized anomaly score, $min(S)$ and $max(S)$ indicate the minimum and maximum losses in the whole dataset.

**Competition-max** $(\tilde{S}_{max})$: using the maximum of multiple normalized anomaly scores to detect anomalies.

$$\tilde{S}_{max}(t) = max\{S'_{int}(t), S'_{grad}(t), S'_{flow}(t), S'_{adv}(t)\} \quad (29)$$

**Competition-min** $(\tilde{S}_{min})$: using the minimum of multiple normalized anomaly scores to detect anomalies.

$$\tilde{S}_{min}(t) = min\{S'_{int}(t), S'_{grad}(t), S'_{flow}(t), S'_{adv}(t)\} \quad (30)$$

**Cascade-normal** $(\tilde{S}_{cn})$: detecting and discarding Cred-n frames in former detectors, and transmitting the remaining frames to the next detector. In the cascade process, $thr^n$ is set as the value calculated in MAAS, $S_{int}$ is set as the last detector.

**Cascade-abnormal** $(\tilde{S}_{ca})$: detecting and discarding Cred-a frames in former detectors, and transmitting the remaining frames to the next detector. In the cascade process, $thr^a$ is set as the value calculated in MAAS, $S_{int}$ is set as the last detector.

The comparison results are shown in table 2.

**Table 2.** AUCs of different aggregation strategies.

| | | Ped2 | Avenue |
|---|---|---|---|
| Raw | $S_{int}$ | 0.9678 | 0.8982 |
| | $S_{flow}$ | 0.98 | 0.8607 |
| | $S_{grad}$ | 0.9689 | 0.842 |
| | $S_{adv}$ | 0.5759 | 0.8441 |
| Weight sum | $\tilde{S}_w$ | 0.9843 | 0.8733 |
| | $\tilde{S}_{wn}$ | 0.9672 | 0.9071 |
| Competition | $\tilde{S}_{min}$ | 0.9123 | 0.8989 |
| | $\tilde{S}_{max}$ | 0.9613 | 0.849 |
| Cascade | $\tilde{S}_{cn}$ | 0.9824 | 0.9027 |
| | $\tilde{S}_{ca}$ | 0.9808 | 0.9124 |
| **Our** | $\tilde{S}_{agg}^{int}$ | **0.9934** | **0.9256** |

From the table we can see that:

(1) $\tilde{S}_w$ performs well in Ped2 dataset, but perform poorly in

Avenue dataset; $\tilde{S}_{wn}$ perform well in Avenue dataset, but perform poorly in Ped2 dataset. We can conclude that the enhancement effect of the weighted sum strategy is not robust. In fact, in many cases, the weighted sum strategy tends to achieve the average performance of multiple detectors. (2) Neither of $\tilde{S}_{max}$ and $\tilde{S}_{min}$ perform well in Avenue and Ped2 datasets. We think the reason is that the anomaly scores of different detectors are not comparable. (3) The cascade strategy (include $\tilde{S}_{cn}$ and $\tilde{S}_{ca}$) can achieve well performance in both datasets. (4) The proposed strategy achieves well performance in both datasets, and achieves the best performance in all strategies. That illstrates the superiority of the MAAS method.

*4.6. Impact analysis*

In fact, MAAS can be regarded as an improved cascade strategy: the master detector plays the similar role of the cascade strategy's last detector; the auxiliary detectors play the similar role of the cascade strategy's former detectors.

Comparing with the cascade strategy, the MAAS has the following advantages: (1) It utilizes both Cred-a frames and Cred-n frames in auxiliary detectors other than using only Cred-n frames. (2) Applying the time continuity property, it extracts more credible information. (3) It utilizes credible information in a voting manner, that makes the credible information more reliable. (4) Utilizing the soft weights to assist the master detector is more effective than believing the credible frames directly. Moreover, unlike cascade strategy detects credible information in series, MAAS detect credible information in auxiliary detectors in parallel. This is helpful to speed up the detection process.

(1) Impact of Cred-a and Cred-n

Table 3 shows the performances of MAAS when using different credible information.

**Table 3.** The frame-level AUCs of MAAS when utilizing different credible information.

|  | Ped2 | avenue |
|---|---|---|
| $\tilde{S}_{agg}^{int}$(Cred_a) | 0.9808 | 0.9221 |
| $\tilde{S}_{agg}^{int}$(Cred_n) | 0.9867 | 0.9043 |
| **$\tilde{S}_{agg}^{int}$(Cred_a & Cred_n)** | **0.9934** | **0.9256** |

It demonstrates that: the more the credible information merged to the aggregated detector, the better performance the aggregated detector can achieve.

(2) Impact of time continuity process

As shown in Fig. 4, by applying time continuity process, we inferred more credible frames. Table 4 exhibits performances of MAAS with and without time continuity process. As shown in table 4, with time continuity process, MAAS achieves better performance.

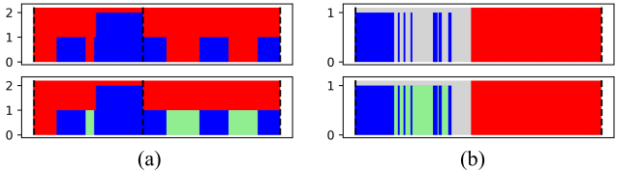

**Fig. 4.** The detected credible frames before and after time continuity process. The red regions indicate abnormal events. The blue vertical lines indicate the detected credible frames. The green vertical lines indicate the inferred credible frames. (a) Cred-a frames. Top: Cred-a frequencies before time continuity process. Bottom: Cred-a frequencies after time continuity process. (b) Cred-n frames. Top: Cred-n frequencies before time continuity process. Bottom: Cred-n frequencies after time continuity process.

(3) Impact of voting mechanism

**Table 4.** Frame-level AUCs of MAAS with and without time continuity process.

|  | Ped2 | avenue |
|---|---|---|
| $\tilde{S}_{agg}^{int}$ without time continuity process | 0.9875 | 0.9154 |
| **$\tilde{S}_{agg}^{int}$ with time continuity process** | **0.9934** | **0.9256** |

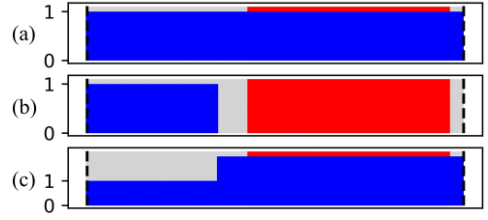

**Fig. 5.** Voting process visualization of a video sample. (a) Cred-a frequencies. (b) Cred-n frequencies. (c) The calculated soft weights.

Fig. 5 visualizes the voting process when calculating soft weights for a video sample.

As shown in Fig. 5, there are misjudgments in Cred-a information. After voting, the Cred-n information corrected these misjudgments.

(4) Effectiveness of soft weight method

When merging the credible information to the master detector, the discarding method can be regarded as a special case of the soft weight method. The discarding method thresholds the soft weights into 3 values:

$$\omega_{soft}(t) = \begin{cases} 0, & if \quad \omega_{soft}(t) < 1 \\ 1, & if \quad \omega_{soft}(t) = 1 \\ +\infty, & if \quad \omega_{soft}(t) > 1 \end{cases} \quad (31)$$

Process of thresholding is a process of losing information. Therefore, the discarding method loses confidence information of the soft weights. Table 5 shows the performances of MAAS

**Table 5.** Frame-level AUCs of MAAS using discard method and soft weight method.

|  | Ped2 | avenue |
|---|---|---|
| $\tilde{S}_{agg}^{int}$(Discard) | 0.9917 | 0.9242 |
| **$\tilde{S}_{agg}^{int}$(Soft weight)** | **0.9934** | **0.9256** |

under soft weight method and discarding method.

As shown in the table, the soft weight method achieves better performance. The reason for this phenomenon is that: the soft weight method allows the master detector to take part in the anomaly detection on the credible frames, which makes the judgements on these credible frames more robust.

*4.7. Comparing with state-of-the-art algorithms*

In table 6, the performance of the proposed strategy is compared with the state-of-the-art algorithms. Applying the proposed strategy, a new state-of-the art performance is achieved.

**Table 6.** Frame-level AUC performances of different methods.

|  | Ped2 | Avenue |
|---|---|---|
| Deep-cascade [16] | 0.939 | N/A |
| WTA-AE [33] | 0.966 | 0.821 |
| U-Net predict [27] | 0.954 | 0.851 |
| STAN [9] | 0.965 | 0.872 |
| Multilevel Representations [31] | 0.9752 | 0.7154 |
| Narrowed cluster [34] | 0.944 | 0.878 |
| $S^2$-VAE [17] | N/A | 0.876 |
| Cross-channel [11] | 0.955 | N/A |
| AnomalyNet [35] | 0.949 | 0.861 |
| sRNN-AE [36] | 0.9221 | 0.8348 |
| U-Net predict reconstruct [25] | 0.963 | 0.851 |
| Block-level-process [29] | 0.9911 | 0.8986 |
| **Our method** | **0.9934** | **0.9256** |

## 5. Conclusion

This paper proposes a new and effective aggregation strategy for video anomaly detection. In this strategy, the auxiliary detectors just need to detect Cred-a and Cred-n frames as accurate as possible. They do not need to detect all anomalies. This reduces auxiliary detectors' burden. The stronger the complementarity among auxiliary detectors, the better performance MAAS can achieve. In the future, we will try to find more complementary detectors and aggregate them.